\documentclass[conference]{IEEEtran}
\IEEEoverridecommandlockouts
\usepackage{cite}
\usepackage{amsmath,amssymb,amsfonts}
\usepackage{algorithmic}
\usepackage{graphicx}
\usepackage{textcomp}
\usepackage{xcolor}
\def\BibTeX{{\rm B\kern-.05em{\sc i\kern-.025em b}\kern-.08em
    T\kern-.1667em\lower.7ex\hbox{E}\kern-.125emX}}
\begin{document}

\title{Fast Object Detection with a Machine Learning Edge Device\\
}

\author{\IEEEauthorblockN{Richard C. Rodriguez, MSDA}
\IEEEauthorblockA{\textit{Information Systems and Cyber Security Department} \\
\textit{The University of Texas at San Antonio}\\
San Antonio, Texas, USA\\
richard.rodriguez@my.utsa.edu}
\and
\IEEEauthorblockN{Jonah Elijah P. Bardos, MS}
\IEEEauthorblockA{\textit{Electrical and Computer Engineering Department} \\
\textit{The University of Texas at San Antonio}\\
San Antonio, Texas, USA\\
jonahelijah.bardos@my.utsa.edu}

}

\maketitle

\color{black}
\begin{abstract}
This machine learning study investigates a low-cost edge device integrated with an embedded system having computer vision and resulting in an improved performance in inferencing time and precision of object detection and classification.

A primary aim of this study focused on reducing inferencing time and low-power consumption and to enable an embedded device of a competition-ready autonomous humanoid robot and to support real-time object recognition, scene understanding, visual navigation, motion planning, and autonomous navigation of the robot.

This study compares processors for inferencing time performance between a central processing unit (CPU), a graphical processing unit (GPU), and a tensor processing unit (TPU). CPUs, GPUs, and TPUs are all processors that can be used for machine learning tasks. Related to the aim of supporting an autonomous humanoid robot, there was an additional effort to observe whether or not there was a significant difference in using a camera having monocular vision versus stereo vision capability. TPU inference time results for this study reflect a 25\% reduction in time over the GPU, and a whopping 87.5\% reduction in inference time compared to the CPU.

Much information in this paper is contributed to the final selection of Google's Coral brand, Edge TPU device. The Arduino Nano 33 BLE Sense Tiny ML Kit was also considered for comparison but due to initial incompatibilities and in the interest of time to complete this study, a decision was made to review the kit in a future experiment.

\end{abstract}

\begin{IEEEkeywords}
CPU, GPU, TPU, inference, AI, machine learning, object detection
\end{IEEEkeywords}

\section{Introduction}
% \color{blue}
Machine learning is a subset of artificial intelligence (AI) that enables computers to extract features from images and learn to recognize patterns. This is important because images can vary widely in quality, lighting, and other factors. AI excels at defining probabilities based on images, which makes it ideal for tasks such as object detection and recognition.

In addition to object detection and recognition, computer vision enables humanoid robots to navigate and interact with the environment. AI provides robots with the ability to perceive their surroundings and understand the spatial relationships between objects. This allows them to detect and track objects, estimate depth, and reconstruct 3D scenes [8].

Use cases for computer vision include recognition of people, faces, vehicles, scene understanding, visual navigation, motion planning, navigation, and other objects. Other applications include medical imaging analysis, industrial inspection, retail analytics, security and surveillance, and autonomous driving.

This study investigates low-cost low-power edge device affixed to an embedded system having a monocular or stereo camera and which utilizes the microprocessor for computational intelligence. One aim is to enable a robot to detect a soccer ball and to act upon the target. This outcome of this study will result in reducing the overall cost of a competitive robot intended to compete in the 2024 International RoboCup competition.

In the use case of computer vision and computational intelligence of autonomous machines or humanoid robots, the primary stakeholders are the creators, owners, or investors. When knowledge of this technology becomes democratized, secondary stakeholders, such as, an audience, fan-base, or the public, in which the technology is deployed may not be as accepting outside of a competition arena [4].

Some ethical ramifications and values of an autonomous machine or robot having computer vision and computational intelligence within or outside the domain of an international robot soccer competitions may include;

Bias and discrimination. Computer vision algorithms are trained on large datasets of images and videos. If these datasets are biased, the algorithms will learn to be biased as well. This could lead to autonomous machines discriminating against certain groups of people, such as minorities, women, or the elderly.

Invasion of privacy. Computer vision systems can be used to track and monitor people's movements without their consent. This raises serious concerns about privacy rights and surveillance.

Lack of informed consent. When people interact with autonomous machines in public spaces, they may not be aware that they are being monitored or that their data is being collected. This lack of informed consent raises concerns about fairness and autonomy.

Security and surveillance concerns. Computer vision systems can be hacked or misused to create surveillance systems that violate people's privacy and security.

Social impact and inequality. The widespread use of autonomous machines in public spaces could have a significant social impact, including job displacement and increased inequality.

We contend there is a moral obligation to ensure that autonomous machines leveraging machine learning, computer vision and computational intelligence, are used in a way that respects our privacy, our rights, and our values.

\begin{figure}[h]
\centerline{\includegraphics[scale=0.25]{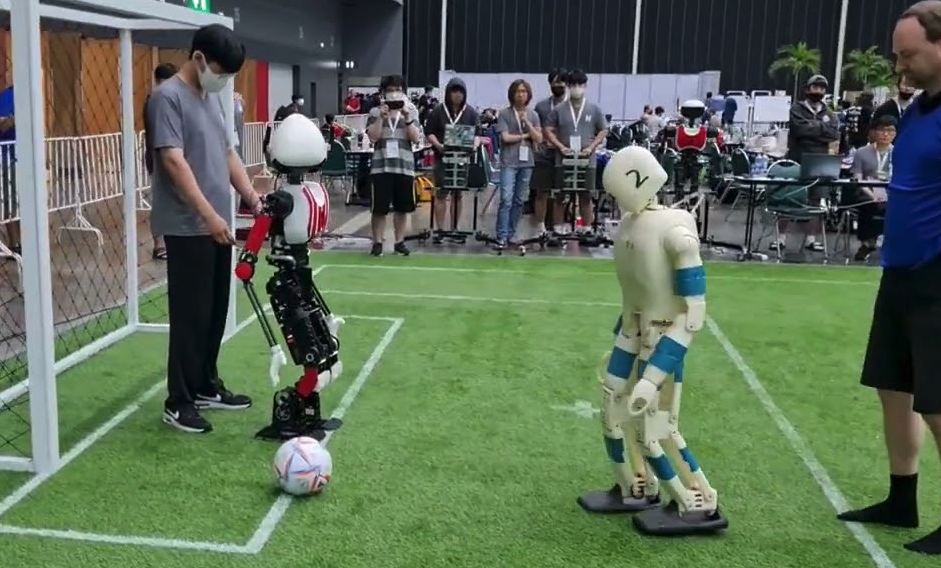}}
\caption{Humanoid Robots at a RoboCup Competition.}
\label{fig:photos}
\end{figure}

\subsection{CPU vs. GPU vs. TPU}
CPUs are general-purpose processors that can be used for many different applications, but they are not as fast as GPUs or TPUs for machine learning tasks. The CPU used in this processor comparison was the 13$^{th}$ Generation Intel Core$^{TM}$ i9-13900H 2.60 GHz.

GPUs are specialized processors that are designed for matrix multiplication, which is a common operation in neural networks. GPUs are faster than CPUs for machine learning tasks, but they are still not as fast as TPUs. The GPU used in this processor comparison was the NVidia brand GEFORCE RTX 4070.

TPUs are specialized processors that are designed for neural network workloads, such as Convolutional Neural Networks (CNNs) appropriately used for computer vision inferencing. TPUs are the fastest type of processor for machine learning tasks. The TPU used in this processor comparison was the Google Coral-brand, Edge TPU.

\begin{table}[h]
\caption{Key Differences between CPUs, GPUs, and TPUs}
\begin{center}

\begin{tabular}{|c|c|c|c|}
\hline

\textbf{}&\multicolumn{3}{|c|}{\textbf{Processor Comparison}} \\
\cline{2-4} 

\textbf{} & \textbf{\textit{CPU}}& \textbf{\textit{GPU}}& \textbf{\textit{TPU}} \\
\hline

General-Purpose Processor & YES & NO & NO  \\
\hline
Speed: ML Tasks & Slowest & Faster & Fastest \\
\hline
Cost / Expense & Middle & Most & Least \\
\hline

\multicolumn{4}{l}{$^{\mathrm{a}}$Coral Edge TPU = {\$31.00} U.S.}
\end{tabular}
\label{tab1}
\end{center}
\end{table}

In general, you should use a CPU for quick prototyping or if you have simple models that do not take long to train. You should use a GPU if you have models for which the source does not exist or is too onerous to change, or if you have models with a significant number of custom TensorFlow operations that must run at least partially on CPUs. You should use a TPU if you have models that are dominated by matrix computations, or if you have models that train for weeks or months at a time.

\section{Hypotheses}

Hypothesis 1: Integrating a Tensor Processing Unit (TPU) with an embedded device or microprocessor with or without a Graphics Processing Unit (GPU) will significantly improve inferencing time and object detection to provide a low-cost and low-power option for embedded devices.

Hypothesis 2: Embedded devices utilizing ML affixed with a monocular camera significantly improves object detection performance at a lower cost compared to stereo vision cameras.

% \color{blue}
\section{Methodology}
\subsection{Convolutional Neural Network}
Figure 2. depicts an overview of a convolutional neural network (CNN) architecture for image recognition [3]. Images from a color camera having three-channel RGB input, perform a convolution + activation function, e.g., Rectified Linear Unit (ReLU) and  max pooling operation repeatedly throughout the convolutional backbone, part of the overall structure of the architecture. From an input image, feature maps are created by convolution and resized by Max Pooling operations getting smaller in resolution but deeper in feature maps until they reach a classifier, the head of the architecture, and at this stage is a fully connected multi-layer artificial neural network.

\begin{figure}[h]
\centerline{\includegraphics[scale=0.28]{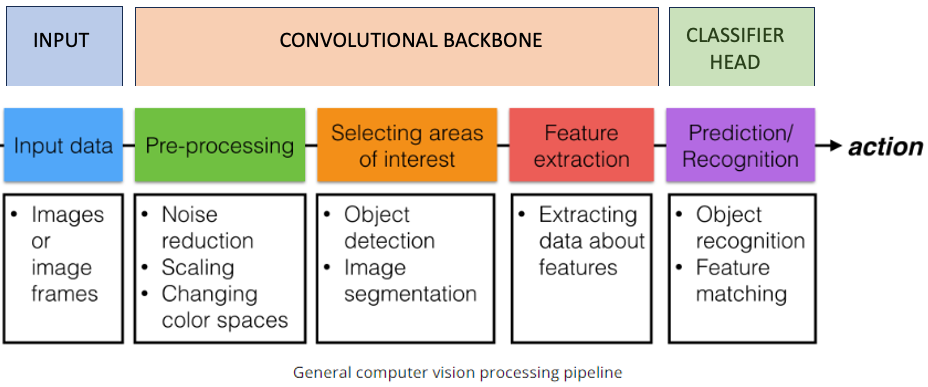}}
\caption{A Convolutional Neural Network Components.}
\label{fig:graphics}
\end{figure}

\color{black}
\subsection{Training and Deployment}

Object detection using convolutional neural networks is a powerful and versatile technique for identifying and locating objects in images. The general process of object detection using CNNs may include the following.

Image Preprocessing: The input image is first preprocessed to ensure consistency and compatibility with the CNN model. This may involve resizing, normalizing, and converting the image to a suitable format.

Feature Extraction: The CNN extracts features from the preprocessed image. These features represent the essential characteristics of the image content and are used to identify and locate objects.

Object Classification and Bounding Box Regression: For each proposal, the CNN classifies whether the proposal contains an object of interest and predicts the bounding box coordinates for the object.

Post-processing: The final output of the object detection system is typically a list of detected objects, each with its corresponding class label, confidence score, and bounding box coordinates.

\color{black}
\section{Hardware}
\subsection{Experiment Kit Options}
Two devices were considered for experimentation, the Arduino Nano 33 BLE Sense ML Kit and the Google Coral Edge TPU device. Due to difficulties with software versions and experiencing difficulties with a headless connection the TPU device was selected as the main device for this study.

% \color{red}
\subsection{Google Coral Edge TPU}
\begin{figure}[h]
\centerline{\includegraphics[scale=0.35, angle=90]{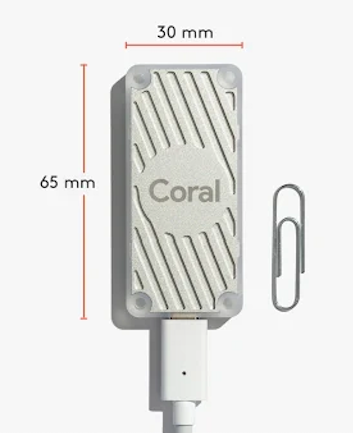}}
\caption{Google Coral TPU.}
\label{fig}
\end{figure}

The Google Coral Edge TPU offered the best inference time performance over the CPU and GPU. Below are the critical hardware components used for this study.

\begin{itemize}
  \item Google Coral Edge TPU
  \item Intel RealsenseTM D35I Stereo Camera
  \item Generic Mono Webcam
  \item Proc: 13th Gen. Intel Core i9-13900H 2.60 GHz
  \item GPU: GEFORCE RTX 4070
\end{itemize}

% \color{red}
\section{Software}
Below is the list of software products to support the hardware used in this study.
\begin{itemize}
  \item Python
  \item OpenCV
  \item YOLOv8
  \item TensorFlow
  \item TensorFlow Lite
  \item Edge TPU Compiler
  \item PyCoral
  \item Visual Studio Code
  \item Anaconda
\end{itemize}

% \color{blue}
\section{Data Source}

The computer vision dataset used as this project's data source originated from was from RoboFlow.com and contained 4,000 images.

\section{Results}
\subsection{Implication of type-1 and type-2 errors}

A type-1 error is a false positive and a type-2 error is a false negative. This is important for stakeholders to understand the ramifications of using ML-based decision and design for the error type that should be avoided the most. It will help choose algorithms based on sensitivity or specificity depending on what the preference is for a use case [4].

\subsection{Precision and Recall}

Precision is the rate at which your detections are most likely going to be true positives. In other words, it is the proportion of positive identifications that are actually correct. For example, if you have a precision of 0.8, then 80\% of the time you identify something as being positive, it will actually be positive.

Recall is the rate at which your detections are correct. In other words, it is the proportion of actual positives that are correctly identified. For example, if you have a recall of 0.9, then 90\% of the time there is something positive, you will correctly identify it.

The formula for precision is True Positive divided by the sum of True Positive and False Positive (P = TP / (TP + FP).

\begin{figure}[h]
\centerline{\includegraphics[scale=0.35]{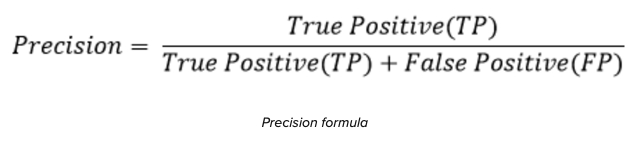}}
\caption{Precision Formula.}
\label{fig1}
\end{figure}

The formula for recall is True Positive divided by the sum of True Positive and False Negative (P = TP / (TP + FN).

\begin{figure}[h]
\centerline{\includegraphics[scale=0.35]{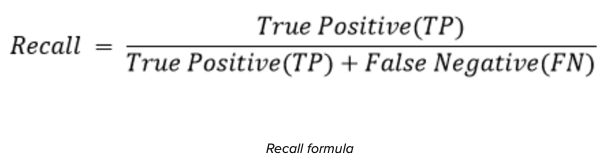}}
\caption{Recall Formula.}
\label{fig2}
\end{figure}

In the case of a detecting a soccer ball for a competitive humanoid robot to use electric energy to activate the kinetic movement of its leg to kick the ball, it is more important to have a high precision rate than a high recall rate. This is because the cost of a false positive (using energy to move the leg when there is no soccer ball) is much higher than the cost of a false negative (not immediately detecting the soccer ball).

Figure 6. reflects a precision rate reaching above 80\% compared to the recall rate of below 80\%. 

\begin{figure}[h]
\centerline{\includegraphics[scale=0.5]{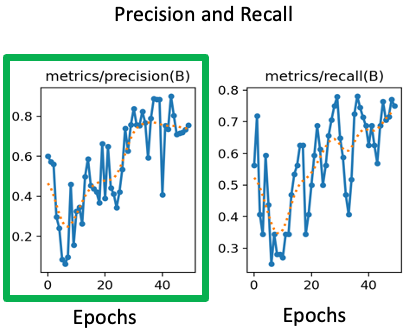}}
\caption{Precision and Recall Curves.}
\label{fig:graphs}
\end{figure}

\begin{figure}[h]
\centerline{\includegraphics[scale=0.45]{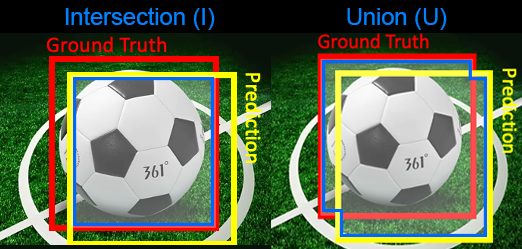}}
\caption{Intersection and Union.}
\label{fig}
\end{figure}

The Intersection over Union (IoU), shown in Figure 7., provides a metric to set this boundary at, measured as the amount of predicted bounding box that overlaps with the ground truth bounding box divided by the total area of both bounding boxes.

\begin{figure}[h]
\centerline{\includegraphics[scale=0.5]{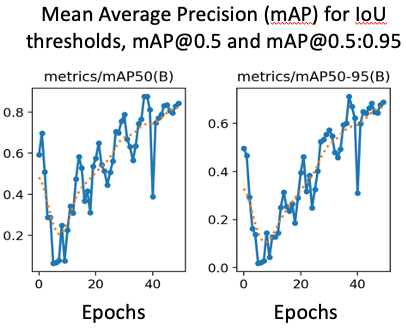}}
\caption{Mean Average Precision (mAP).}
\label{fig3}
\end{figure}

\begin{figure}[h]
\centerline{\includegraphics[scale=0.4]{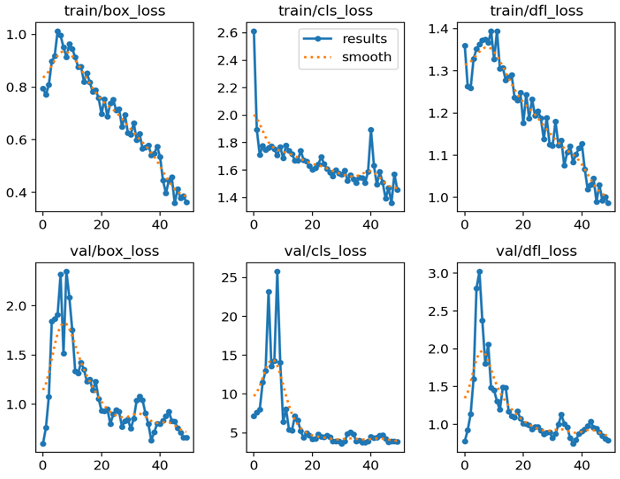}}
\caption{Mean Average Precision (mAP).}
\label{fig3}
\end{figure}

% \color{red}
\subsection{Training and Validation}

The program was scripted to run 50 epochs until the training process delivers acceptable levels of accuracy, Figure 8. and Figre 9. Separate training and test data at an 80/20 split ensured the neural network did not accidentally train on data used later for evaluation. We took  advantage of transfer learning or utilized a pre-trained network and repurposed it. TensorFlow was used to help improve accuracy and minimize box-loss, classification loss (CLS), and dual focus loss (DFL).

% \color{red}
\subsection{Inference Time Processor Performance}
Results on inference time performance for each processor, CPU, GPU, and TPU are depicted in Table II, below. 

\begin{table}[h]
\caption{Live Inference Time Performance}
\begin{center}
\begin{tabular}{|c|c|c|c|}
\hline
General-Purpose Processor & CPU & GPU & TPU  \\
\hline
Real-Time Inference & 240 ms & 40 ms & 30 ms \\
\hline
% \multicolumn{4}{l}{$^{\mathrm{a}}$Coral Edge TPU = {\$31.00} U.S.}
\end{tabular}
\label{tab1}
\end{center}
\end{table}

% \color{blue}
\section{Conclusion and Future Work}
Our study concludes having a powerful CPU does not offer a significant advantage for object detection. Integrating an ML inferencing edge device (Edge TPU) with embedded devices demonstrates a viable option to operate at low-power and minimum cost. Additionally, there was no significant difference in using a mono vision camera over stereo vision.

Future work will be integrating the Coral Edge TPU device with the on-board processor of competitive humanoid robot to detect soccer balls.

\subsection{GitHub Code Link}
The  GitHub repository link for Richard Rodriguez and Jonah Elijah Bardos for this project and programming code is noted below.
\[{https://github.com/ThinkFastAI/AI\_Practicum} \]

% \[{------------------}\]

% \color{blue}

\end{document}